\documentclass[letterpaper, 10 pt, conference]{ieeeconf}  % Comment this line out if you need a4paper

\IEEEoverridecommandlockouts                              % This command is only needed if 
\usepackage{graphics} % for pdf, bitmapped graphics files
\usepackage{amsmath} % assumes amsmath package installed
\usepackage{amssymb}  % assumes amsmath package installed
\usepackage{todonotes}
\usepackage{physics} % for \abs{} symbol
\usepackage{algorithm}
\usepackage[noend]{algpseudocode}
\usepackage{xcolor}

\title{\LARGE \bf
Online Object Model Reconstruction and Reuse\\ for Lifelong Improvement of Robot Manipulation}

\author{Shiyang Lu$^{1}$, Rui Wang$^{1}$, Yinglong Miao$^{1}$, Chaitanya Mitash$^{1}$, Kostas Bekris$^{1}$% <-this % stops a space
\thanks{$^{1}$The authors are affiliated with the Department of Computer Science at Rutgers, the State University of New Jersey, New Brunswick, NJ, 08901, USA. Email: shiyang.lu@rutgers.edu; kostas.bekris@cs.rutgers.edu
        }%
\thanks{This work is partially supported by NSF NRI award IIS 1734492 and HDR TRIPODS award 1934924.}
}

\begin{document}

\maketitle
\thispagestyle{empty}
\pagestyle{empty}

%%%%%%%%%%%%%%%%%%%%%%%%%%%%%%%%%%%%%%%%%%%%%%%%%%%%%%%%%%%%%%%%%%%%%%%%%%%%%%%%

\begin{figure*}[t]
    \centering
    \includegraphics[width=0.85\textwidth]{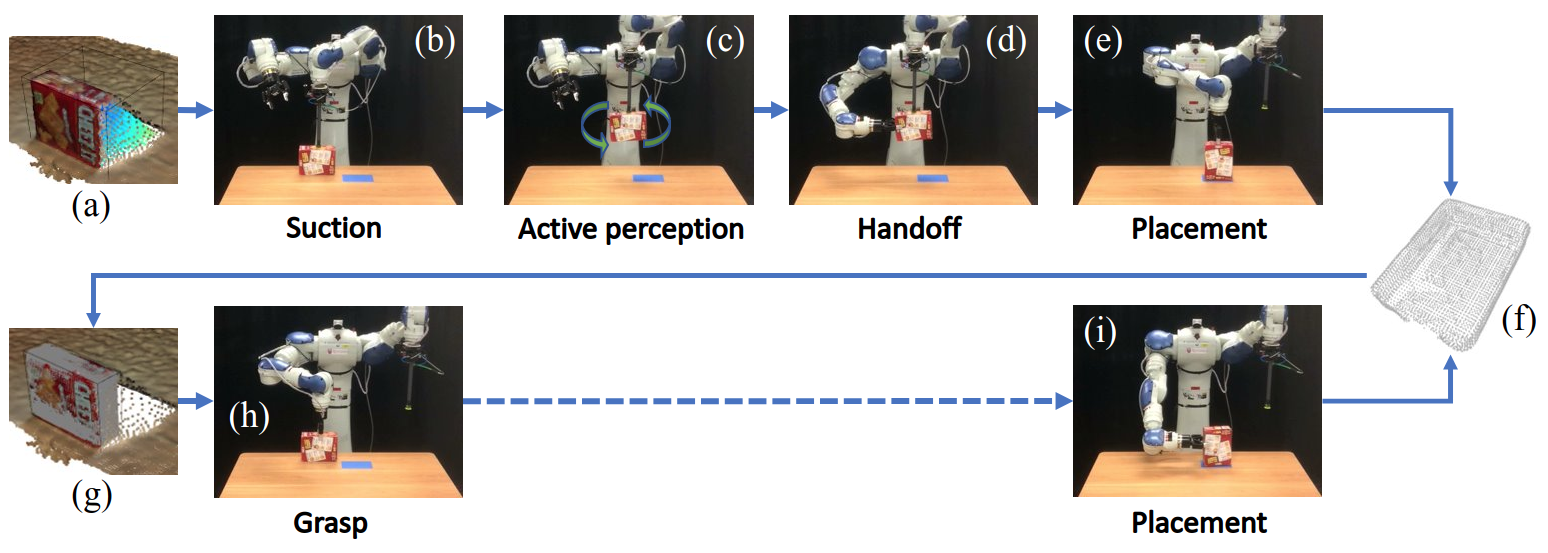}
    \vspace{-.1in}
    \caption{(top row) A manipulation sequence for an unknown object observed for the first time. Model reconstruction is executed on the fly to get an object model. (bottom row) The same object is recognized and its reconstructed model registered against the observation resulting in a more efficient manipulation sequence. The task involves picking (images b and h), and placing the object (e and i). Without a known geometric model, this task may not be directly solvable and may require multiple intermediate actions, such as \textit{active perception} (c), or reorientation through actions, such as \textit{handoffs} (d). Reconstructing and reusing object models during manipulation allows to avoid these actions in future episodes due to reduced uncertainty. In this lifelong process, the object models (f), are continuously updated over time given new viewpoints.} \vspace{-.1in}
    \label{fig:motivation}
\end{figure*}

\begin{abstract}
This work proposes a robotic pipeline for picking and constrained placement of objects without geometric shape priors. Compared to recent efforts developed for similar tasks, where every object was assumed to be novel, the proposed system recognizes previously manipulated objects and performs online model reconstruction and reuse. Over a lifelong manipulation process, the system keeps learning features of objects it has interacted with and updates their reconstructed models. Whenever an instance of a previously manipulated object reappears, the system aims to first recognize it and then register its previously reconstructed model given the current observation. This step greatly reduces object shape uncertainty allowing the system to even reason for parts of objects, which are currently not observable. This also results in better manipulation efficiency as it reduces the need for active perception of the target object during manipulation. To get a reusable reconstructed model, the proposed pipeline adopts: i) TSDF  for object representation, and ii) a variant of the standard particle filter algorithm for pose estimation and tracking of the partial object model. Furthermore, an effective way to construct and maintain a dataset of manipulated objects is presented. A sequence of real-world manipulation experiments is performed. They show how future manipulation tasks become more effective and efficient by reusing reconstructed models of previously manipulated objects, which were generated during their prior manipulation, instead of treating objects as novel every time.

% This work aims on supporting the lifelong manipulation of objects, which are originally unknown in terms of 3D models at the instance level. Using a recent pipeline for pick-and-constrained placement without geometric models, it shows how to construct models effectively online during this unknown object manipulation. The resulting object models are then used during the future manipulation of the same object instance. This is achieved by recognizing that object from an actively maintained dataset. The object model is then registered against the currently observed RGBD data. This work uses a sequence of real-world manipulation experiments to show how much more effective future manipulation tasks can be executed by reusing such reconstructed models generated on the fly instead of treating objects as novel every time. The model reconstruction process, however, needs to be accurate enough to support effective recognition so as to get these benefits in manipulation. For this purpose, this work proposes a particle-filter based approach for simultaneously reconstructing the model and tracking the object while it is being manipulated. The same approach is used for model matching in future manipulation episodes. Furthermore, an efficient way is proposed for constructing a dataset to store color and geometric information of manipulated objects based on a TSDF volume representation and feature vectors of cropped images. 
\end{abstract}

%%%%%%%%%%%%%%%%%%%%%%%%%%%%%%%%%%%%%%%%%%%%%%%%%%%%%%%%%%%%%%%%%%%%%%%%%%%%%%%%
\section{Introduction}
General purpose and flexible robot manipulators should be able to manipulate object instances they have never seen before. Once an object has been manipulated, however, the robot should be able to leverage its prior experience for future encounters of a similr object. Such abilities allow the deployment of robots that self-learn to precisely manipulate objects and improve their performance over time.

Recent work on manipulating novel objects either completes their shape based on category-level reasoning \cite{gao2019kpam, gualtierirobotic} or utilizes physical constraints \cite{agnew2020amodal}. Single-view 3D shape completion, however, is often not precise and safe enough for many manipulation tasks, such as pick and place in constrained spaces. To ensure safety during manipulation of novel objects,  recent prior work \cite{mitash2020task} considers a conservative estimate of an object's volume. The conservative estimate includes the observed surface of that object together with the volume attached to it, which has not yet been observed bounded by physical constraints, such as the presence of a support surface. To achieve this, the prior work uses a simple volumetric object representation similar to occupancy-grids \cite{occ_grid}, and proposes action primitives, such as \textit{handoffs} and \textit{active perception}, to reduce shape uncertainty of novel objects during manipulation. While this prior system can deal with constrained placement tasks for novel objects, its success rate and efficiency sometimes suffers. This is due to the fact that every object is treated as novel, even if instances of the same object have been seen before. The current work aims to improve the efficiency of such manipulation pipelines by recognizing previously manipulated objects and reusing the models that it has constructed online given prior manipulation operations. During the development of the proposed work, it was identified that occupancy-grid representations are not precise enough for reusing the reconstructed object model in future tasks.

To address these issues, this paper proposes a  manipulation pipeline that performs object picking and constrained placement via life-long object model reconstruction and reuse. It is based on the hypothesis that some objects will reappear over multiple manipulation tasks. For instance, this can occur in logistics setups where singulated objects are dropped from a conveyor belt and then need to be picked and placed in a container to be shipped. To achieve more accurate reconstruction and model reuse, the Truncated Signed Distance Function (TSDF) is adopted to represent partial models. Similarly, a variant of a standard particle filter \cite{6696485, 10.1145/3240765.3243493, chen2019grip} is used for performing pose estimation and tracking of the partial TSDF models. This variant prunes pose hypotheses that violate viewpoint or physical constraints, and rejects models of falsely recognized objects from being reused. It achieves speed advantages by rendering objects in a region of interest instead of a full image. Furthermore, this work presents an effective way to construct a dataset on the fly that stores partially reconstructed object models for future tasks. This dataset also stores a set of color features for each object that are the output of a clustering algorithm given previous object viewpoints. The clustering is experimentally shown to result in efficient and accurate object recognition.

A sequence of real-world experiments with a manipulator and a set of objects is performed to show that more successful and more efficient robot manipulation can be achieved over time by proper reconstruction and reuse of object models. Compared to the baseline \cite{mitash2020task}, the proposed robotic system not only achieves higher success rate (by a 13\% margin), but also significantly improves efficiency. In particular, it reduces \textit{handoff} actions by 31\%, and reduces \textit{active perception} actions by 49\% over the same sequence of manipulation tasks against the baseline.

\section{Related Work}
\subsection{Pick-and-place Manipulation for Novel Objects}
Robot manipulation systems for tight packing \cite{shome2019towards} or placement in constrained spaces \cite{haustein2019object} often assume the availability of complete 3d object models. For novel object instances, a majority of recent work focuses on task-agnostic picking \cite{mahler2019learning, zeng2018robotic} while others resort to shape completion, performed either via category-level reasoning \cite{gao2019kpam, gualtierirobotic} or given physical consistency constraints \cite{agnew2020amodal}. Nevertheless, the output of shape completion may not be precise enough and lead to collisions when the object is placed in a constrained space. Recent work \cite{mitash2020task} proposes to use a conservative shape representation for pre-pick planning to ensure safe manipulation and reconstruct the shape of the object in-hand, if the task requires it. In certain scenarios, the conservative estimate might be too restrictive for the constrained placement task. To address this, the current work proposes to recognize previously seen objects and perform life-long model reconstruction over many manipulation runs. 
\vspace{-0.05in}

\subsection{Simultaneous Tracking and Object Reconstruction}
Object models are often generated by using a turntable \cite{calli2015yale}, or via manual scanning  \cite{wang2019hand} or a robotic arm \cite{krainin2010manipulator} and post-processed. These models are then used for single-shot pose estimation \cite{mitash2018robust, sui2017sum} or model-based object tracking \cite{choi2012robust}. Model-free manipulation research has used local scan matching \cite{icp} with an occupancy grid structure \cite{mitash2020task} to simultaneously track and reconstruct a conservative object volume. Another popular surface
reconstruction technique often used in SLAM is Truncated Signed Distance Function (TSDF) \cite{curless1996volumetric, newcombe2011kinectfusion}. It fuses multiple depth observations from a moving sensor and maintains a signed distance to the closest zero-crossing (representing the surface). The current work leverages  TSDF in a particle filter for simultaneous tracking and reconstructing a manipulated object.
\vspace{-0.05in}

\subsection{Object Identification and Pose Estimation}
Previous work \cite{sharif2014cnn, babenko2014neural} has shown that features trained on large-scale classification datasets allows for image matching. The current work leverages such pre-trained features to store object viewpoints and re-identify object instances. Pose estimation based on particle filters \cite{sui2017sum} has been used before for matching complete object models with object segments in the scene. The current work utilizes a similar technique but with partial object models that were constructed from past manipulations. 

\begin{figure*}[h]
    \centering
    \includegraphics[width=\textwidth]{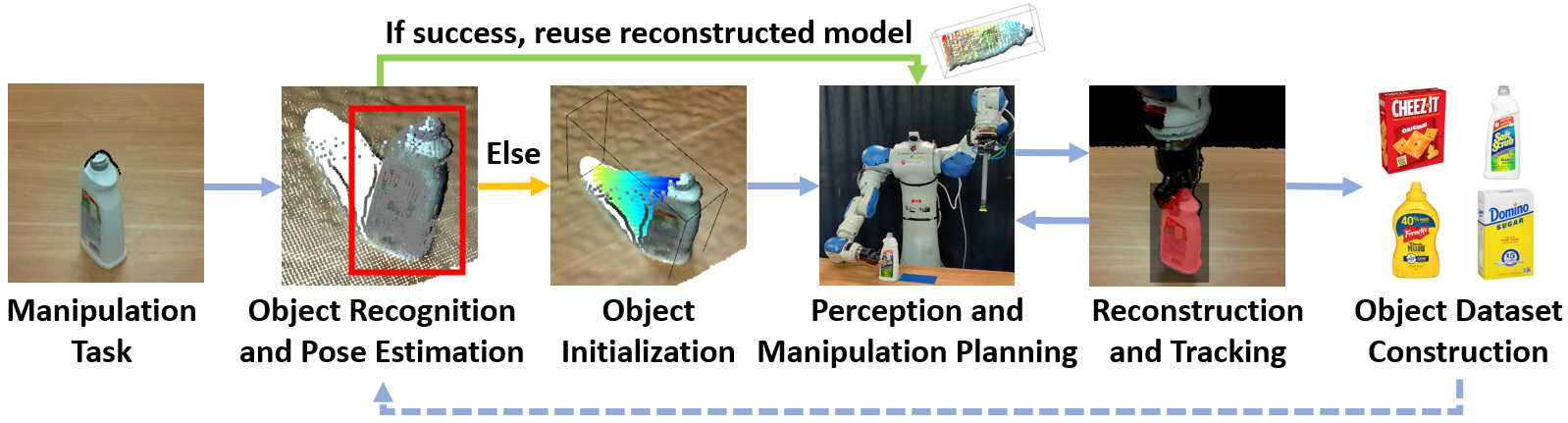}
    \vspace{-.25in}
    \caption{Proposed pipeline: For each object, recognition is first executed given an object dataset. If the object is recognized, pose estimation given its existing (perhaps partial) model is performed and the model is reused. If not, a new, partial model is initialized from  current data. Either way, a perception and manipulation process is executed. If the model is not completed enough, additional sensing and reorientation action may be needed to place the object in a constrained area. Visual tracking is executed to dynamically update the object model in parallel. The latest model is stored in the dataset.}
    \label{fig:pipeline}
    \vspace{-.2in}
\end{figure*}

\section{Problem Setup and Notation}

\textbf{Object Representation}
The $i^{th}$ rigid object to be manipulated is defined by the volume it occupies $O^i \in \mathbb{R}^3$ in a local reference frame. Given a pose $P^i \in SE(3)$, the 3D region occupied by $O^i$ in the global frame is denoted as $O^i_{P^i}$. Note that the ground truth geometric model of this object is not available, i.e., $O^i$ is unknown. The estimated object representation $\hat{O}^i$ is composed of the object's surface $S^i$ and a conservative volume $U^i$, which has not yet been viewed but may contain part of the target object, i.e. $\hat{O}^i = S^i \cup U^i$.

\textbf{Object Recognition and Pose Estimation:} One singulated object $O^i$  appears for picking for each task $i$. The robot first determines whether $O^i$ has been manipulated before given the initial RGB-D observation $I^i_{init}$. If $O^i$ is recognized as in the same category as $O^j$, where $j < i$, then the reconstructed model $\hat{O}^j$ is reused to initialize $\hat{O}^i$, which is the registered output of the current object point cloud in $I^i_{init}$ and $\hat{O}^j$. The initial pose $P^i_{init}$ is set to be $\hat{P}^i$, which is the estimated pose of the object model $\hat{O}^i$ given $I^i_{init}$. If $O^i$ is recognized as a novel object, then $\hat{O}^i$ is directly initialized from the object's point cloud in $I^i_{init}$.

\textbf{Constrained placement}
Given an object at a pose $P^i_{init}$, the goal of the constrained placement is to move $O^i$ to a pose $P^i_{target}$, such that $O^i_{P_{target}} \subset R^i_{place}$ where $R^i_{place} \in \mathbb{R}^3$ is a target placement region. To accomplish this, a sequence of manipulation actions, i.e. \textit{pick}, \textit{handoff}, \textit{place} are performed. Since the ground truth $O^i$ is unknown, perceptive actions that sense the object may also be needed. 

%%%%%%%%%%%%%%%%%%%%%%%%%%%%%%%%%%%%%%%%%%%%%%%%%%%%%%%%%%%%%%%%%%%%%%%%%%%%%%%%
\section{System Design and Implementation}
The proposed pipeline is shown in Fig. \ref{fig:pipeline}. At the beginning of each task, the robot first determines whether the target object has been seen before. If an object is considered novel, then its model will be initialized based on the current RGB-D observation. Otherwise, a previously reconstructed model, which may be partially complete, is registered to the current observation and reused in the current task. An integrated perception and manipulation planning process is performed thereafter to accomplish a constrained placement task. During manipulation, the object is tracked and its model is dynamically updated. The reconstructed model is also used by the manipulation planning process, which forms a feedback loop. A dataset, which stores object information, is updated after each manipulation task to benefit future tasks.

\smallskip
\subsection{Object Initialization} An object model is initialized when the target object is considered unknown. Then, a truncated signed distance function (TSDF) \cite{tsdf} representation is used as the object model. TSDF has been widely used for high-quality scene reconstruction \cite{kinectfusion, bundlefusion}. Each voxel in a TSDF volume stores the signed distance $d$ to its closest surface, where the sign of $d$ indicates whether the voxel is in free space $(d > 0)$ or in a conservative estimate of the object's volume $(d < 0)$. The surface point cloud $S^i$ of the object can be extracted at the zero crossings of the function either by ray-casting or a marching cubes algorithm \cite{lorensen1987marching}. An object's TSDF volume is initialized given a minimum oriented 3D bounding box that encloses the observed point cloud of an object and its occluded region. Standard methods are used to approximately compute this 3D bounding box \cite{malandain2002,barequet2001}. The point cloud segment of the object can be easily obtained since each task only contains one object. The voxel size of a TSDF volume is set to be 1mm in the accompanying implementation, which is small enough to capture object details for both 3D registration and grasp pose detection.

%  The conservative volume $U^i$ of the object (i.e., the volume either inside the object or the volume not observed so far) corresponds to the region occupied by voxels with $d < 0$. 

\subsection{Dataset Construction} \label{dataset_construction}
A dataset is constructed from scratch to store information of manipulated objects. For each object instance, a reconstructed TSDF model and a set of RGB features are stored. Cropped RGB images captured during manipulation are fed to a neural network for extracting features representing the object. The implementation uses ResNet50 \cite{He_2016_CVPR} pretrained on ImageNet \cite{5206848} as the feature extractor similar to related work \cite{8461044}. The choice of the specific feature extractor is not the main focus of this work and can be replaced with alternatives. Since an object may look very different from different viewpoints, a set of features vectors are computed via clustering to represent an object instead of a single feature vector. In particular, the mini-batch KMeans algorithm that has been designed for efficient incremental clustering of new data \cite{sculley2010web, scikit-learn}, is adopted to cluster features from similar viewpoints. It is called after each manipulation task. Given an ablation study (Sec. \ref{obj_recognition}), a value of $K=64$ for the number of clusters used for an object's features provides good viewpoint diversity and near instant fitting speed. 

% \begin{figure}[h]
%     \centering
%     \includegraphics[width=0.45\textwidth]{example-image-b}
%     \caption{Object Recognition and Pose Estimation}
%     \label{fig:estimation}
% \end{figure}

\begin{figure}[h]
\vspace{-.1in}
    \centering
    \includegraphics[width=0.48\textwidth]{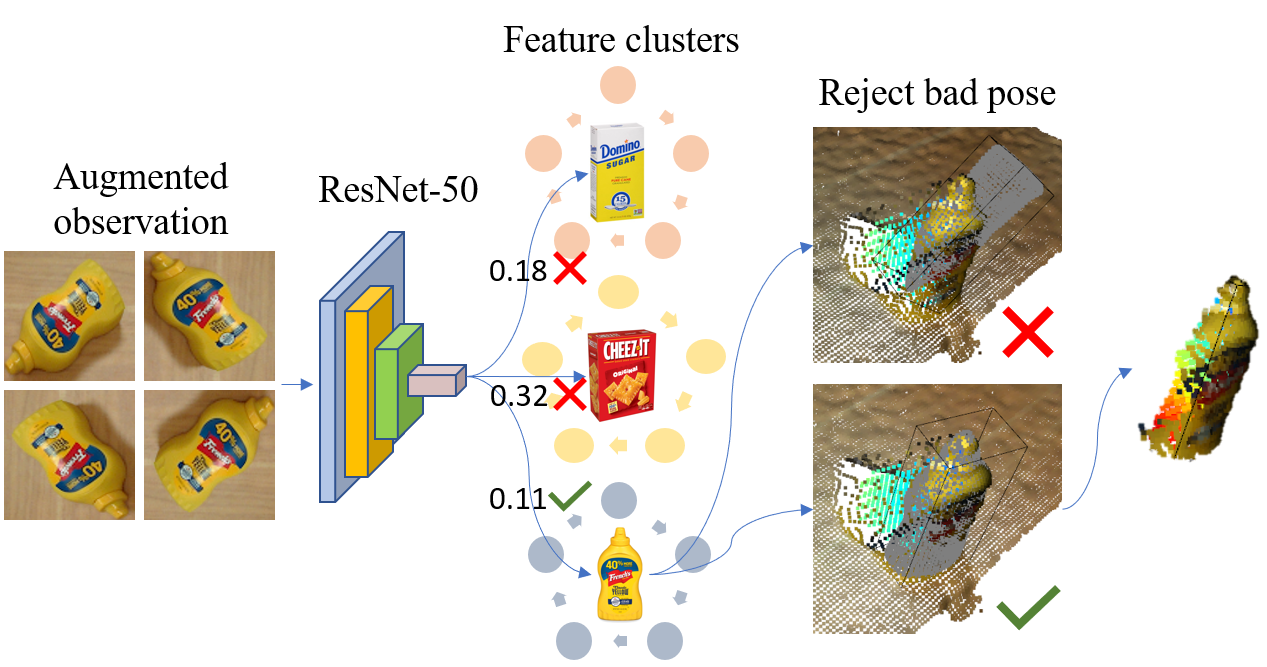}
    \vspace{-.25in}
    \caption{Object Recognition and Pose Estimation. A segmented observation is first augmented by rotation and then fed to a neural network for feature extraction. The features are compared with centers of feature clusters of objects in the dataset. The object with the closest cosine distance that is less than a threshold ($\delta = 0.15$) is considered as a matching candidate. Pose estimation with viewpoint constraints is performed to reject false positives.}
    \label{fig:estimation}
    \vspace{-.2in}
\end{figure}

\subsection{Object Recognition and Pose Estimation}
\label{sec:objrecposeest}
Given an initial observation $I^i_{init}$ for task $i$, the robot first attempts to recognize $O^i$ and estimate its pose $P^i_{init}$. Despite notable progress in object recognition and 6D pose estimation, these problems remain challenging in the considered setup, since: 1) the dataset is constructed from scratch and data collected from one task can be insufficient to train a deep model; 2) retraining a deep model after each task is time consuming and violates the objective of performing efficient manipulation. This work proposes a two-stage method that performs object recognition and pose estimation without retraining a feature extractor multiple time as in Fig. \ref{fig:estimation}.

First, a cropped observation is augmented rotation-wise 8 times and fed to a feature extractor. The features for each rotated image are then compared using cosine distance against the K cluster centers of the feature sets selected to represent each object. Among the nearest neighbors of all rotated images within a cosine distance $d < \delta = 0.15$, the most similar nearest neighbor is selected as a matching candidate $\bar{O}^i = \hat{O}^j$. If no nearest neighbor has a distance $d < \delta$, then this object is considered to be novel. An ablation study of the threshold $\delta$ is performed in section \ref{obj_recognition}.

\vspace{-.1in}
\begin{algorithm}[h!]
\caption{6DoF Pose Estimation}\label{alg:pf}
\begin{algorithmic}[1]
\Require Observed depth image $I$, object TSDF volume $V_{obj}$, extrinsic matrix $E$, intrinsic matrix $C$, number of particles $M$, number of iterations $K$, pixel depth inlier threshold $d_{thres}$, rejection ratios $\beta_1, \beta_2$.

\noindent
\State Generate scene TSDF volume $V_{scene}$ from $I$, $E$, and $C$.
\State Filter table and get object region of interest for rendering.
\State Compute 3D centroid $c$ of $I_{roi}$ projected in 3D space.
\State Initialize a set of $M$ particles at $t=0$, $\mathcal{X} = \{x^1_t, ..., x^M_t\}$ located at position $c$ with random orientation in $SO(3)$.
\For{$t = 1$ to $K$}
    \State $\mathcal{\hat{X}}_t = \mathcal{X}_t = \emptyset$
    \For{$m = 1$ to $M$}
        \State Diffuse $x^m_t \sim p(x_t | u_t, x^m_{t-1})$ \Comment{$u_t$ is zero.} \label{ln:diffuse}
        \State Render object depth $I_r$ in RoI given $V_{obj}$ and $x^m_t$
        \State $w^m_t =$ count\_pix\_inliers($I_{roi}$, $I_{r}$, $d_{thres}=1cm$)
        \State $\mathcal{\hat{X}}_t = \mathcal{\hat{X}}_t$ + $\langle x^m_t, w^m_w \rangle$
    \EndFor
    \For{$m = 1$ to $M$}
        \State Draw $x^i_t$ with probability $\propto$ $w^i_t$
        \State $\mathcal{X}_t = \mathcal{X}_t + x^i_t$
    \EndFor
\EndFor
\State Sort $\mathcal{X}_K$ based on particle weights (descending order)
\State $x_{best} \gets$ None
\For{$m = 1$ to $M$}
    \State Render object $I_r$ at $x^m_K$ and project to $V_{scene}$
    \State $\beta_{free} \gets$ pts\_ratio\_in\_freespace($I_r, V_{scene}, C, E$)
    \State $\beta_{collide} \gets$ pts\_ratio\_below\_table($I_r, V_{scene}, C, E$)
    \If{$\beta_{free} < \beta_1$ and $\beta_{collide} < \beta_2$}
        \State $x_{best} \gets x^m_K$
        \State \textbf{break}
    \EndIf
\EndFor
\State \textbf{Return} $x_{best}$
\end{algorithmic}
\end{algorithm}
\vspace{-.1in}

Then, a particle filter variant is used to estimate the pose of the object candidate $\bar{O}^i$, which can help reject potential false positives in recognition. The variant is detailed in Alg. \ref{alg:pf} and is adapted from existing Monte Carlo localization methods \cite{6696485, 10.1145/3240765.3243493, chen2019grip}, with the following differences: 1) It can work on partially reconstructed TSDF volumes other than complete mesh models; 2) Rendering is only performed in a Region of Interest (RoI - referred to $I_{roi}$ in the algorithmic), which is the smallest 2D bounding box of the conservative volume estimate augmented by a $30\%$ margin. This makes the algorithm more efficient ($\sim30ms/iter$ vs $\sim60ms/iter$ \cite{chen2019grip}) with the same number of particles ($N=625$); 3) Two rejection criteria are introduced to prune bad pose hypotheses that either violate viewpoint constraints or physical constraints, i.e. a registered model should not lie in the free space or collide with the supporting plane. If the number of pixel violating these criteria over the total number of pixels in the ROI is less than $\beta_1$ and $\beta_2$ respectively, where $\beta_1$ and $\beta_2$ are predefined thresholds ($\beta_1 = \beta_2 = 0.95$ in the accompanying implementation), then it is considered a good registration. Otherwise, $O^i$ is considered novel and will be initialized from scratch. These additional pose rejection criteria minimize the chance of a falsely recognized object to be registered and fail a manipulation task. The algorithm is implemented using PyCUDA \cite{kloeckner_pycuda_2012}, and the authors have open sourced on Github as a 6DoF pose annotation tool.

\subsection{Simultaneous Reconstruction and Tracking} \label{tracking}
The object model $\hat{O}^i$ is tracked and reconstructed over time. The same particle filter variant as in Alg. \ref{alg:pf} is reused with the following changes: 1) The object transition model (in line \ref{ln:diffuse}) between two time steps is set to be: $u_t = \Delta E^i_{t-1:t}$, instead of 0, where $E^i_t$ is the end-effector's pose computed via forward kinematics; 2) One iteration is performed for each new observation; 3) The RoI is computed by first rendering the object at $\hat{x} = u_t \cdot \hat{x}_{t-1}$, where $\hat{x}_{t-1}$ is the most likely estimate, and then finding its minimum 2D bounding box augmented by a 30\% size increase; 4) The rejection criteria are not used for tracking. New observations are then integrated to the object's TSDF volume after the arm and the end-effector are filtered. Since the ground-truth, frame-to-frame tracking pose of a model under reconstruction is not available, tracking quality is implicitly evaluated by the final object reconstruction (Sec. \ref{obj_reconsturction}).

\begin{figure}[h]
\vspace{-.15in}
    \centering
    \includegraphics[width=0.48\textwidth]{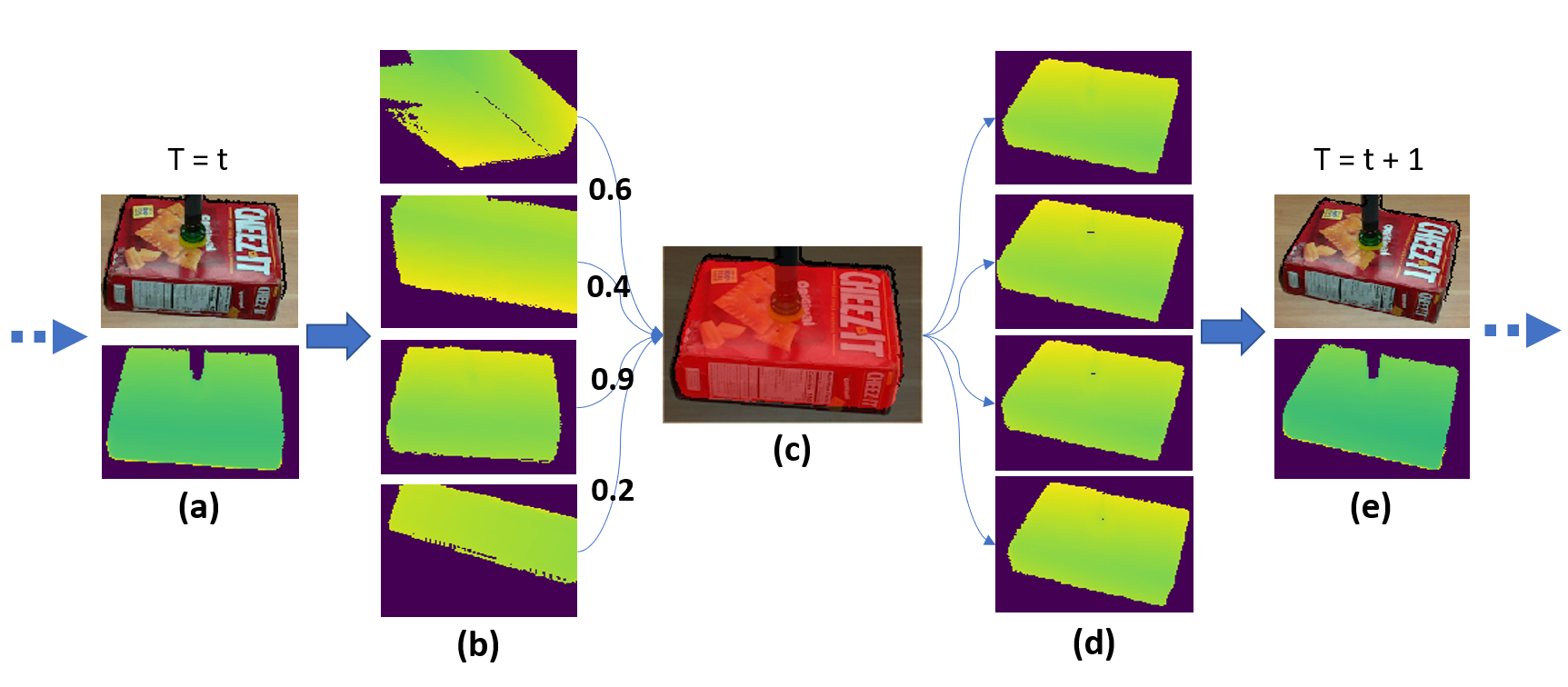}
    \vspace{-.3in}
    \caption{Illustration of Simultaneous reconstruction and tracking. Given a new observation at $T=t$ (a), a transition model given a diffusion process is applied on particle estimates, which are sampled given the previous time step, and rendered at the region of interest (b). The particle with the largest weight is selected as the pose prediction at $T=t$ (c). The current observation is integrated to the object representation and particles are then resampled (d) to become available for the next time step (e).}
    \label{fig:tracking}
    \vspace{-.15in}
\end{figure}

\subsection{Perception and Manipulation Planning}
Given a specified constrained area $R^i_{place}$, the robot will first check if the target object can be directly grasped by its end-effectors given the object's conservative volume. The dual-arm robot prioritizes the two-finger gripper as the orientation of the suction gripper is limited. Grasp poses are computed over the surface representation $\hat{S}^i$ to ensure stable geometric interaction \cite{ten2017grasp}. For the suction gripper, suction points are sampled on the object surface $\hat{S}^i$, where the normalized normal $N$ is close to the global $z$ axis i.e., $\abs{N_z} > 0.8$. Suction points are further ranked in quality according to their distances from the center of the surface.

Given the constrained area $R^i_{place}$ and the object representation $\hat{O}^i$, two bounding boxes are computed: 1) the maximum bounding box inside the constrained area, i.e., $B^i_{place} \subset R^i_{place}$, 2) a minimal bounding box $B_{O^i}$ that encloses $\hat{O}^i$. A discrete set of configurations (= 24) for the object's bounding box are computed by placing $B_{O^i}$ at the center of $B^i_{place}$ and are validated by all axis-aligned rotations. If no placement can be found, an active perception action will be taken to move the object to the front of the camera and rotate along the z-axis by $180$ deg. to reduce model uncertainty, and recompute the object bounding box and placement. If there exists a placement but the target pose is beyond the reachability limits of the suction gripper, a \textit{handoff} action will transfer the object from the suction to the parallel gripper. 

%The bounding boxes as well as the place poses will be recomputed.

% Given a target pose, a planning module is need to compute an executable trajectory which moves the robot arm to a specified target pose without introducing collisions of the robot arm and the manipulated object with regard to the environment. It takes as the input estimated object representation and a set of target gripper poses, and returns a sequence of prehensile manipulation actions. 
% The planning is categorized into two modes (1) transit planning, which refers to the planning of robot motions to approach and grasp the object. It loops through all candidate grasp poses until a grasp pose is both reachable and collision-free. Then a pre-grasp pose is generated which is 10cm behind the chosen grasp pose in the approaching direction (local $Z$-axis of the end effector frame). The planner first generates a sequence of motions which moves the arm to the pre-grasp pose and then approach to the specified grasp pose with cartesian move and (2) transfer planning, which refers to the planning of robot motions to transfer the object in hand to a designated pose. Through the process of in-hand manipulation, object tracking is performed and new object information will be obtained. In the absence of any errors, the execution of these actions solves the constrained placement task. 

As part of the planning framework, a probabilistic roadmap ({\tt PRM}$^*$) \cite{kavraki1996probabilistic, Karaman:2011aa} is pre-computed for each of the arms, which takes into account collisions with static obstacles, such as the table. To generate informative paths for the considered setup, the configurations along the roadmap are sampled so that the end-effector is within the camera's view. This allows the object to be tracked during arm movement. Based on the precomputed {\tt PRM}$^*$ roadmap, a lazy version of the {\tt A}$^*$ algorithm is used online for computing a shortest path on the roadmap, where lazy collision checks with the object are performed after an initial solution path is found. Once a collision has been detected, the roadmap is modified and a new {\tt A}$^*$ query is triggered. The loop continues until a valid path is confirmed.

\section{Experiments}
Experiments are designed to showcase the effectiveness and efficiency of the proposed robotic manipulation system. It is compared with a baseline system \cite{mitash2020task}, which was designed to perform similar manipulation on unknown objects but considers all objects as novel without learning object information from tasks or reuse reconstructed object models in future tasks. In addiition, evaluation and ablation studies are performed to show the efficacy of the system's submodules. In particular, the proposed system is evaluated from the following perspectives: 1) Success rate, 2) system efficiency, 3) reduced shape uncertainty after registration, 4) object recognition, and 5) object reconstruction.

\subsection{Hardware and Experimental Setup}
\begin{figure}[ht]
    \centering
    \includegraphics[width=0.4\textwidth, height=5cm]{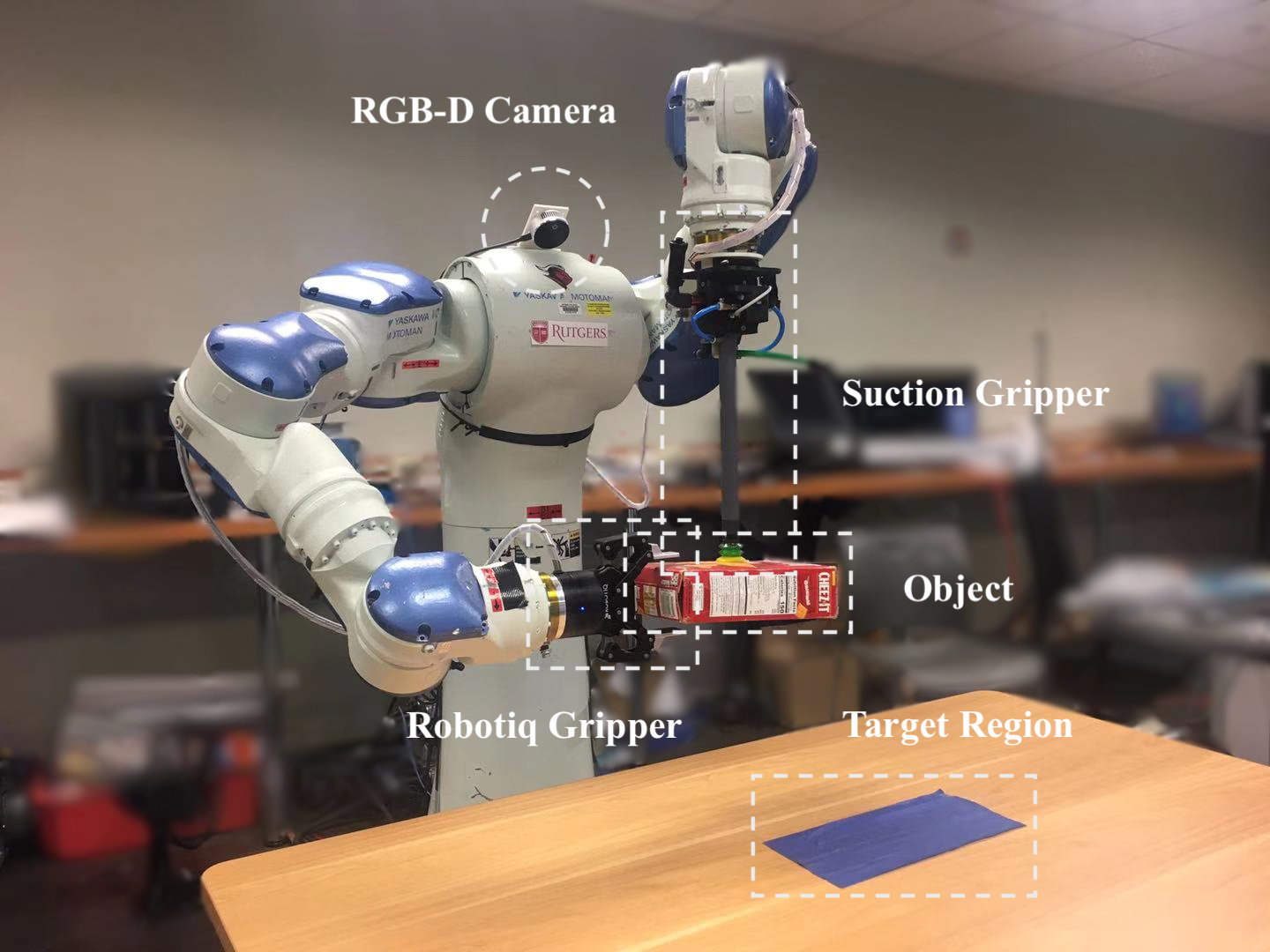}
    % \vspace{-.1in}
    \caption{The hardware setup involving a dual-arm manipulator with a suction and a parallel gripper manipulating objects in the presence of a static RDG-D camera so as to achieve placement in a target region.}
    \label{fig:hardware}
    % \vspace{-.1in}
    \vspace{-.2in}
\end{figure}

The hardware setup is shown in Fig. \ref{fig:hardware}. It comprises of a dual-arm manipulator (Yaskawa Motoman) with two 7-dof arms. The left arm is fitted with a suction gripper, while the right arm is fitted with a Robotiq 2-fingered gripper. A single RGB-D sensor (RealSense L515) is mounted on the robot torso overlooking the workspace. The camera is configured to capture 480p RGB-D images at a frequency of 30 Hz. 

%In addition to real world experiments, we also do simulation experiments with more objects in PyBullet \cite{coumans2019} with the exact same robot setup except for the camera model.

Randomly ordered constrained placement tasks are performed given 4 objects (shown in Fig. \ref{fig:reconstruction_qualitative}). Each task requires the robot to pick up an object from the table and place it in a constrained area. The target object is randomly positioned on the 2D plane within the reach of both grippers. The objects are placed on different sides and rotated along the z-axis by a predefined angle for each experiment. Since the \textit{bleach} object can't be placed stably on its side but either stand or lie flat on the table, only 10 experiments were performed for it. For \textit{cheezit}, \textit{sugar}, and \textit{mustard}, 15 experiments are performed for each. In total, 110 real world pick and place experiments were executed for both systems for comparison.

\subsection{Results}
%  Due to lack of ground truth pose for partially reconstructed objects, it is difficult to report quantitative results for the accuracy of pose estimation. Nevertheless, the visualization output indicated high accuracy in all of the experiments performed.

\subsubsection{Success Rate}
There are certain object configurations that correspond to a large initial conservative estimate of an object, and thus no safe grasps or top-down suction points can be found, e.g., a standing bleach cleanser. In these cases, the task will fail if uncertainty is not reduced. In a sequence of 55 constrained placement experiments, eight tasks failed for the baseline experiment, while only one task failed for the proposed experiment, which is due to a standing \textit{mustard} not being recognized. Table \ref{table:1} provides the statistics. 

{\renewcommand{\arraystretch}{1.2}%
\vspace{-.1in}
\begin{table}[h]
\centering
\begin{tabular}{|c|c|c|c|} 
\hline
            & {\bf Success Rate} & \# Handoffs & \# Active Perception \\
\hline
Baseline \cite{mitash2020task} & 47/55(85\%) & 37/55(67\%) & 35/55(63\%) \\
\hline
Proposed & 54/55(98\%) & 20/55(36\%) & 8/55(14\%) \\
\hline
\end{tabular}
\caption{Task success rate (higher is better)\\ and ratio of primitive actions used (lower is better)}
\label{table:1}
\vspace{-.3in}
\end{table}

}

% \vspace{-.1in}
% \begin{table}[h]
% \centering
% \begin{tabular}{|c|c|c|c|} 
% \hline
%             & {\bf success rate} & \# handoff & \# active perception \\
% \hline
% baseline \cite{mitash2020task} & 47/55(85\%) & 37/55(67\%) & 35/55(63\%) \\
% \hline
% proposed & 54/55(98\%) & 20/55(36\%) & 8/55(14\%) \\
% \hline
% \end{tabular}
% \caption{Task success rate (higher is better)\\ and ratio of primitive actions used (lower is better)}
% \label{table:1}
% \vspace{-.3in}
% \end{table}

\subsubsection{System Efficiency}
This is evaluated by counting the number of times two action primitives are used, i.e. \textit{active perception} and \textit{handoff}. \textit{Active perception} means that the robot moves the object in front of the camera and rotates it along $Z$ to reduce shape uncertainty. \textit{Handoff} means the robot transfers the object between grippers to achieve a grasp that allows placement. Both actions can be potentially avoided given a better model. The fewer times these actions are taken, the more efficient the manipulation process is. 

%For the baseline method, 37/55 of the tasks require \textit{active perception} actions, and 35/55 of the tasks require \textit{handoff} actions. While for the proposed system where object models are reconstructed and reused, the portion of tasks that take \textit{active perception} and \textit{handoff} actions are greatly reduced to 8/55 and 20/55 respectively.

% \begin{figure}[h]
% \vspace{-.1in}
% \centering
%     \includegraphics[width=0.48\textwidth]{figrues/experiments_stats.png}
%     \vspace{-.3in}
% \caption{Efficiency and failure rates over a sequence of 55 experiments.}
%     \label{fig:running_time_results}
%     \vspace{-.1in}
% \end{figure}

\subsubsection{Reduced Shape Uncertainty}
\begin{figure}[h]
\vspace{-.1in}
    \centering
    \includegraphics[width=0.45\textwidth]{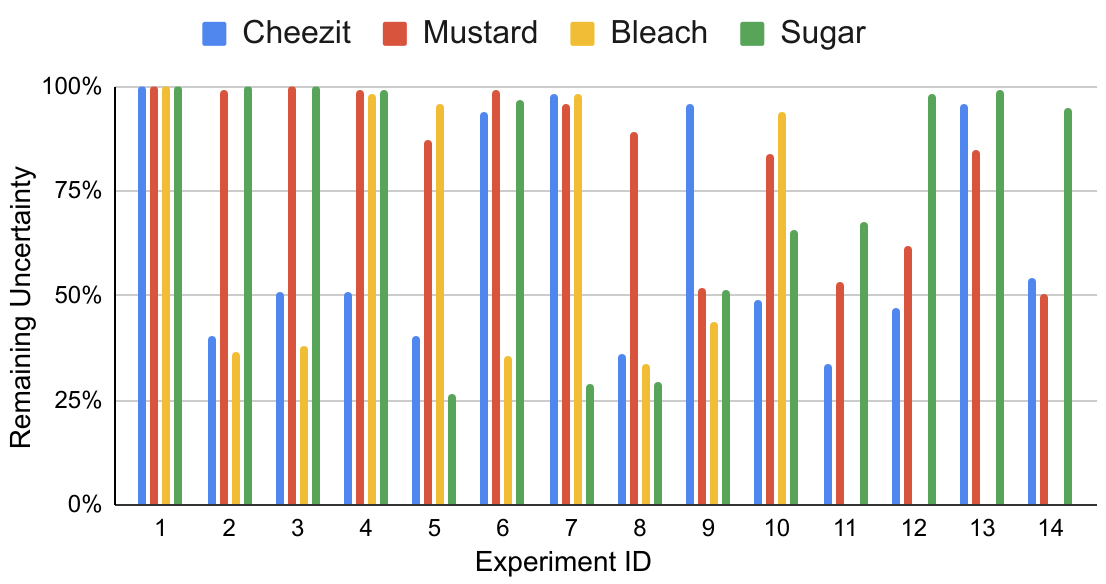}    \vspace{-.15in}
    \caption{Remaining uncertainty of each experiment.}
    \label{fig:reduced_uncertainty}
        \vspace{-.15in}
\end{figure}

By registering a previously reconstructed model, the initial conservative estimate of an object is greatly reduced. This allows a robot to detect more potential grasps and increases the collision-free space for motion planning. As a reminder, the conservative estimate of an object's volume is defined as the ground truth volume of that object together with the volume attached to it, which has not yet been observed. Then, the  shape uncertainty is defined to be the ratio of the conservative estimate of an object's volume after a model has been registered over the conservative estimate of that object given only the current observation. The result for each experiment is shown in Fig. \ref{fig:reduced_uncertainty}. This smaller this ratio is, the more uncertainty is reduced by registering against a previously constructed model. This ratio is reduced by $32\%$ on average, and up to $75\%$ in some cases for all the experiments by reusing a partially reconstructed model.

\subsubsection{Object Recognition} \label{obj_recognition}
The target object is correctly recognized in $49/55$ of the real experiments. In $5/55$ of the experiments the object is erroneously not recognized as a previously manipulated object (false negative). Only one object was initially falsely recognized as a previously manipulated object (false positive), but was then rejected in the pose estimation stage. A small cosine distance threshold $\delta = 0.15$ is used for feature matching to minimize false positives (as in Sec. \ref{sec:objrecposeest}). And a value $K = 64$ is used for the mini-batch K-Means clustering approach (as in Sec. \ref{dataset_construction}), which represents the features of each manipulated object from different viewpoints. False negatives typically occur when the currently observed object part has not be seen in previous experiments. While false positives may cause task failures, false negatives only decrease efficiency as the object will be considered novel. To make the results more statistically meaningful, the data collected from this sequence of experiments were shuffled and an ablation study was performed for the values of the $\delta$ and $K$ parameters. Table \ref{table:2} shows how these two values affect precision and recall. The results are computed from an average of 30 randomly shuffled sequences.

{\renewcommand{\arraystretch}{1.2}%

\vspace{-.1in}
\begin{table}[h]
\centering
\begin{tabular}{|c|c|c|c|c|} 
\hline
            & K = 1 & K = 16 & K = 64 & K = 128 \\
\hline
$\delta$=0.14 & 0.995/0.131 & 0.975/0.755 & 0.981/0.831 & 0.988/0.844 \\
\hline
$\delta$=0.15 & 0.996/0.211 & 0.955/0.842 & 0.958/0.886 & 0.964/0.891 \\
\hline
$\delta$=0.16 & 0.978/0.315 & 0.920/0.892 & 0.932/0.927 & 0.940/0.923 \\
\hline
$\delta$=0.17 & 0.931/0.484 & 0.848/0.936 & 0.862/0.961 & 0.844/0.964 \\
\hline
\end{tabular}
\caption{Ablation study of $\delta$ and $K$. Each cell shows the precision/recall of the recognition.}
\label{table:2}
\vspace{-.3in}
\end{table}
}

Table \ref{table:2} shows that a smaller $\delta$ tends to increase precision but decrease recall. A good balance is achieved when precision is high $(>0.95)$, while keeping recall at a good level for manipulation efficiency. Simply using the mean feature (K = 1) to represent an object is not ideal as the recall is very low. Increasing K is often beneficial, but the benefit diminishes when K becomes very large (e.g., K=128), while also increasing training time increases. 

% Results of object recognition are shown in Fig. 7, which counts the number of times an object that has been previously manipulated is correctly recognized (true positive), not recognized (false negative), and falsely recognized (false positive).

% \vspace{-.1in}
% \begin{figure}[h]
%     \centering
%     \includegraphics[width=0.48\textwidth]{figrues/recognition_results.png}
%     \vspace{-.2in}
%     \caption{Recognition results over a sequence of 55 experiments.}
%     \vspace{-.1in}
%     \label{fig:recognition_results}
% \end{figure}

\subsubsection{Object Reconstruction} \label{obj_reconsturction}
\begin{figure}[h]
\vspace{-.1in}
    \centering
    \includegraphics[width=0.48\textwidth]{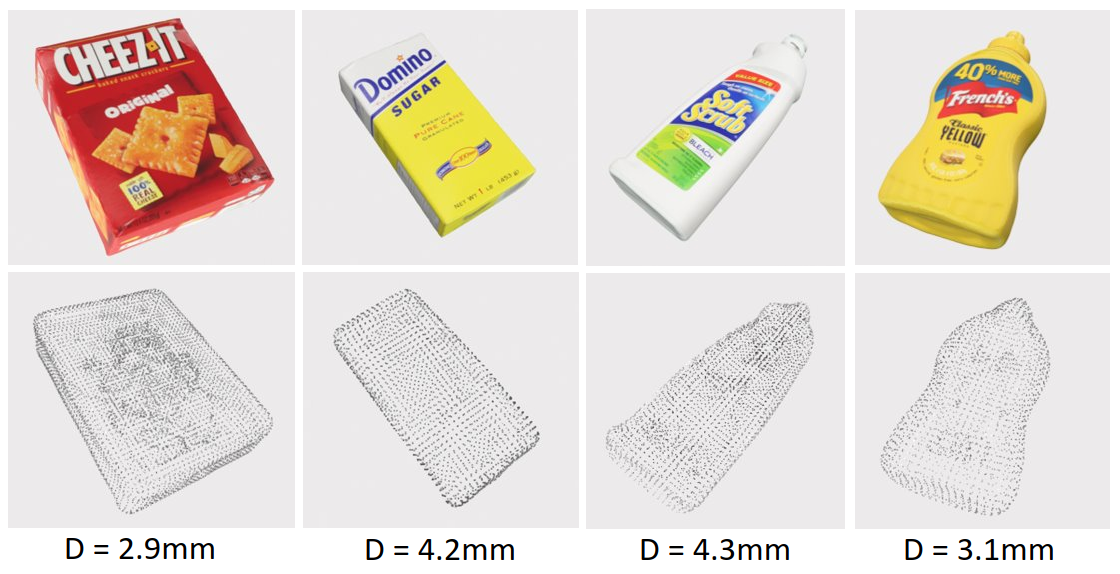}
\vspace{-.25in}
    \caption{Qualitative results of object reconstruction after 55 experiments. Ground truth models are shown in the first row and reconstructed models are shown in the second row. $D$ is the Chamfer distance between the ground truth model and the reconstructed model.}
\vspace{-.1in}
    \label{fig:reconstruction_qualitative}
\end{figure}

Results of shape reconstruction are shown in Fig. \ref{fig:reconstruction_qualitative}. The first row shows the ground truth mesh models of objects, and the second row shows reconstructed models after a sequence of manipulation tasks. For objects that are stored multiple times in the dataset due to failures in recognition, this figure only shows the most completed one. Fig. \ref{fig:reconstruction_qualitative} also presents the quantitative evaluation by comparing the distance between the aligned ground truth model $P_{gt}$ and the reconstructed model $P_{rec}$ using Chamfer distance, i.e.
\vspace{-0.05in}
$$
D(P_{gt}, P_{rec}) = \frac{1}{|P_{gt}|}\sum_{p_i \in P_{gt}}d(p_i, p_r),
\vspace{-0.05in}
$$
where $p_r$ is the closest point in $P_{rec}$ to $p_i$. It can be seen that the reconstructed model is close to the ground truth (Chamfer distance $D < 5mm$ for all four objects) with a few noisy points inside. Such noise is caused by tracking errors when an object is highly occluded, but it does not affect tracking or pose estimation since it will not be rendered by ray casting, and can be further removed by post-processing. 

%%%%%%%%%%%%%%%%%%%%%%%%%%%%%%%%%%%%%%%%%%%%%%%%%%%%%%%%%%%%%%%%%%%%%%%%%%%%%%%%
\section{CONCLUSION}
This work proposes a robotic system, which utilizes object model reconstruction and reuse for achieving lifelong robot manipulation. By using TSDF representations of objects and a particle filter approach for simultaneous reconstruction and tracking, object models are incrementally reconstructed over a sequence of manipulation tasks. An efficient object dataset construction is proposed to store the color and geometry information of manipulated objects, which makes models reusable and assists future manipulation tasks. Real world experiments show the efficiency of the proposed pipeline. 

While this pipeline has been designed to work for most novel rigid objects, it faces challenges with certain objects, such as bowls, which have thin surfaces. This is mainly because the TSDF representation is not suitable for reconstructing thin structures, and may be improved by considering alternatives. Future work will focus on manipulation tasks in cluttered scenes, improving tracking and reconstruction for objects with thin surfaces, and task planning that maximizes information gain while placing novel objects. % But it can also be generalized to scenes with multiple objects given a segmentation model for unknown objects, such as   \cite{lccp,xie2020unseen}. 


%%%%%%%%%%%%%%%%%%%%%%%%%%%%%%%%%%%%%%%%%%%%%%%%%%%%%%%%%%%%%%%%%%%%%%%%%%%%%%%%
\bibliographystyle{IEEEtran}
\bibliography{reference}

\addtolength{\textheight}{-12cm}

\end{document}